# Deduction over Mixed-Level Logic Representations for Text Passage Retrieval

Michael Hess
University of Zurich, Dept. of Computer Sciences
CH-8057 Zurich, Switzerland
`hess@ifi.unizh.ch`

**Abstract**

*A system is described that uses a mixed-level representation of (part of) the meaning of natural language documents (based on standard Horn Clause Logic) and a variable-depth search strategy that distinguishes between the different levels of abstraction in the knowledge representation to locate specific passages in the documents. Mixed-level representations as well as variable-depth search strategies are applicable in fields outside that of NLP.*

## 1. Outline

Most knowledge representation schemes used in AI (and, in particular, in Natural Language Processing) are homogeneous. One application that makes obvious the limitations of a homogeneous representation space is fact retrieval from natural language texts. Our understanding of the semantics of natural language is still so incomplete that the representation of the content of natural language texts must either remain fragmentary, or allow for expressions of varying degrees of abstractness to occur in the same representation scheme. We present, in the context of passage retrieval, a mixed-level representation scheme based on standard Horn Clause Logic and outline a variable-depth evaluation strategy to be used over this type of representation. The combination of such a multi-level representation and a search strategy that is sensitive to the resulting variability in granularity is of general interest for the design of knowledge representation schemes.

## 2. The Problem

The amount of textual information available today in machine-readable form makes it increasingly difficult to locate relevant information reliably and efficiently. With the number of machine readable documents accessible in contemporary data bases or over the networks going into the millions, even the best search systems based on traditional Information Retrieval (IR) methods overwhelm the user, in many cases, with thousands of documents. It would be extremely useful to have text based *fact retrieval systems* or, even more ambitiously, text based *question answering systems*. However, modeling the deep understanding of unrestricted text needed for these applications is still beyond our technical capabilites. This is why we are developing, as an intermediate solution, a system that is capable of taking us to the exact place in a natural language text that is relevant to a query (i.e. a passage retrieval system).

Work on this paradigm has just begun, mainly by researchers in the field of IR ([1 ]). However, it is doubtful that IR methods will be very useful in this new context due to their lack of precision. One of the reasons for this lack lies in the fact that these systems ignore almost all the linguistically relevant information in documents beyond the bare lexical skeleton. Thus *both* of the questions

1)   ?- Logic for natural language analysis

2)   ?- Languages for the analysis of natural logic

would be converted, under a keyword-based approach, into a query

```
?- logic & natural & language & analysis
```

and hence return the same documents although the original queries are far from synonymous. By ignoring function words, morphology, and word order, the keyword extraction procedure loses all the syntactic structure in the natural language input, and consequently all the semantic information that is encoded in natural language through syntax becomes unavailable. Any amount of statistical sophistication applied to the bare keywords will not be able to recover the information that was thrown away when extracting the keywords. This effect makes itself felt particularly clearly when one tries to use standard IR techniques to the goal of passage retrieval.

The goal of this paper is to show how a combination of well known techniques from Natural Language Processing (NLP) and AI, together with the less widespread techniques of mixed-level knowledge representation and variable-depth evaluation strategies, can overcome, in part, the performance related problems that would otherwise mar a fully-fledged NLP based passage retrieval system.

## 3. The use of linguistic information in Information Retrieval so far

Most of the IR systems described in the literature that make use of linguistic information at all fall into the category of syntax driven automatic indexing. In such systems, syntax structures are merely used to derive phrase descriptors, i.e. multi-term descriptors which are then used in the standard way. However, retrieval results are not significantly better than with standard approaches ([2]). Some systems now use the syntactic structures *themselves* as descriptors. In such systems, the syntax structure of a query will have to match the syntax structure of some expressions in a candidate documents *directly*, i.e. without the intermediary step of phrase descriptors (e.g. [3], [4]).

However, if we use full syntax structures as descriptors, we run into the problem that there will almost never be an exact match between the syntax structure of query and (parts of the) documents. First, the same state of affairs can almost always be described in a variety of ways and, in particular, widely different syntactic structures can be used to express the same meaning. For a direct syntax matching approach this has the consequence that we can not even hope to find an *immediate* match between the syntax structure of the query and the syntax structure of (part of) a candidate document. Second, queries normally denote *supersets* of what the relevant documents denote. If our query is ''programming languages'' we want to retrieve (among others) all the documents about ''*object-oriented* programming languages''. Again, the syntax structures of the query and of the expressions making up the documents will be different, and we need some kind of syntactic correlate of the semantic superset/subset relation.

## 4. Using logic as search and index language

An alternative solution that was implemented in a prototype passage retrieval system, LogDoc, starts from the idea that the problem underlying syntactic variability is really a mismatch between a syntactic structure and a semantic expression: Although their *syntactic realisations* are different, the *meanings* of two phrases may be the same, and it is *this* relationship that we are ultimately interested in. The basic idea of LogDoc is, thus, to use logic, with a simple ontology, as a knowledge representation language. We translate documents into logical axioms with back-pointers to the source text, add the axioms to an incrementally growing logical data base, translate queries into theorems, and by proving the latter over the former we are able to retrieve the original documents answering the query. - Key to the success of this procedure is the fact that we can express in logic certain complex relationships between word senses that we cannot represent with standard IR representation schemes. If we use First Order Logic as representation language a document fragment like 3

3) **A structure sharing representation of language for unification based grammar formalisms**

might, for instance, become

```
3a) representation(R,L) ∧ language(L) ∧ share(R,S)
    ∧ structure(S,Y) ∧ goal(F,R) ∧ formalism(F,G) ∧
    grammar(G,Z) ∧ unification(U) ∧ base(F,U)
```

Note that a large number of nouns in English and related languages are of the *relational* type, i.e. they denote relationships. One standard way to represent them in logic is by means of predicates of the appropriate arity, i.e. the same way we will normally represent verbs. Thus two-place predicates like `representation(R,L)`, `formalism(F,G)` etc. express the facts that we are talking about the representation *of* languages, formalisms *for* grammars etc. Note that none of these relations can be represented by the standard IR operators (such as '*adjacent*' etc.) since one of the properties of natural language is that the distance between functionally linked words can (often) be *arbitrarily* long.

The expression 3a is, however, not a logical sentence, i.e. it has no truth value and cannot, as such, be treated as an axiom. But in a retrieval context it is justifiable to say that, whatever is referred to by means of a full noun phrase, is asserted to exist. We can therefore translate everything that is explicitly referred to in a text by means of a full noun phrase into an existentially quantified statement, i.e. we can apply *''existential closure''* to formulae like 3a and get:

```
3b) ∃ R,L,S,Y,F,G,Z,U:
    representation(R,L) ∧ language(L) ∧ share(R,S) ∧
    structure(S,Y) ∧ goal(F,R) ∧ formalism(F,G) ∧
    grammar(G,Z) ∧ unification(U) ∧ base(F,U)
```

In order to have back-pointers to the original documents we add, to each individual axiom, document number and, for passage retrieval, fragment (i.e. sentence, title, or caption) number as additional non-logical constants. If we use a subset of First Order Logic for which efficient proof procedures are known, such as Horn Clause Logic, we will get

```
3c)  representation(sk-1,sk-2)/1/3, language(sk-2)/1/3,
     share(sk-1,sk-3)/1/3, structure(sk-3,sk-4)/1/3,
     goal(sk-5,sk-1)/1/3, formalism(sk-5,sk-6)/1/3,
     grammar(sk-6,sk-7)/1/3, unification(sk-8)/1/3,
     base(sk-5,sk-8)/1/3.
```

meaning that these axioms were all derived from fragment number *1* in document number *3*. Passage retrieval can now be interpreted as proving queries over the logical data base derived from the documents. If we apply a standard proof technique like refutation resolution, a query must be translated into the clausal form of its negation, and

### 4) ?- Structure sharing representations of languages

would then become

```
4a) ?- representation(R,L)/S/D, language(L)/S/D,
       share(R,S)/S/D, structure(S,Y)/S/D.
```

Note, first, that this query means that we want to prove the theorem over axioms that were all derived from the *same* sentence in the *same* document. Note also that the case where a query is more specific than a relevant document (the normal case) is taken care of automatically by this proof procedure, i.e. theorem 4a can be proved *directly* over the axiom system 3c.

## 5. The choice of a suitable ontology

One point that previous approaches to the problem of formalising the semantics of natural language have paid insufficient attention to is that of the *ontology*, or conceptualisation. The question of what kind of *objects*, and what kind of *relationships* between these objects, we assume to exist in the world has a direct bearing on the model to use, and therefore on the definition of the language. The first problem to address is that of representing relationships in logic. There are two basic approaches to this: The ''ordered argument'' approach, and the ''thematic role'' approach (cf. [5 ]:84ff.]). The ordered argument approach assumes that there exists a finite number of roles that objects can play in relationships. These roles are *implicitly encoded* by means of the fixed argument positions of predicates. A sentence like

### 5) John gave Mary an apple

might thus be represented as

```
5a) gave(john,sk-1,mary). apple(sk-1).
```

This is the approach taken in all the examples so far. However, there are many more roles expressed in natural language than those used so far, and for all of them we need to create additional argument positions. Take example

### 6) On Tuesday, John furtively gave Mary an apple in the courtyard

where we express information about the circumstances of the event, viz. about the *time*, the *location* and the *manner* in which the action was performed. We would thus have to extend the predicate pattern by additional argument positions and write

```
7) give(john,sk-1,mary,tuesday1,courtyard1,furtive).
   apple(sk-1).
```

In natural language, there is a rich set of other circumstantial modifiers, for the *cause* and/or for the *reason* of an event ('...*because* he loves her'), for the goal/purpose (''in order to impress her''), or even for highly complex roles such as a concessive circumstance ('...*although* he was told not to'). It seems entirely unclear how many such roles ought to be expressed. There is growing consensus that it is not even a finite set.

Since the ordered argument model seems to fail to account for these empirical facts of language the alternative account, the thematic role account, has gained much popularity. This approach assumes a potentially infinite set of event roles. They are *explicitly encoded* as predicate names. The corresponding predicates range over one object each, and the event itself. What was an *n*-place predicate under the ordered argument account thus becomes a set of *n+1* two-place predicates. 6 would yield a logical representation like

```
7a) eventuality(give,sk-5). time(tuesday1,sk-5).
    agent(john,sk-5). location(courtyard1,sk-5).
    aff_ent(sk-1,sk-5). manner(furtive,sk-5).
    goal(mary,sk-5). object(sk-1,apple).
```

where the Skolem constant **sk-5** denotes the eventuality (the term ''eventuality'' subsumes actions, events, and states). If additional information concerning other event roles becomes available, the set of predicate names is monotonically extended (`reason`, `concessive` etc.), that is, we need never commit ourselves to a fixed set of roles.

It is sometimes assumed that a commitment for one of these two approaches is a matter of technical convenience. It is worth pointing out that this is not the case.

Consider, for instance, the case of action modification (expressed mainly through manner adverbials): By translating 6 as 7 we implicitly assert, by the very act of creating argument positions for time and place, that these attributes are equally fundamental to a goal-directed action as are agent, affected entity, and goal. In particular, by creating an argument position we are forced, under the Horn Clause Logic model assumed here, to specify that the corresponding variable is either universally or existentially quantified. We cannot leave its quantificational status unspecified. However, in natural language we seem to allow unspecified attributive values. While utterance 8 is syntactically well-formed (no time and place is specified), 9 is not (no affected entity is specified):

**8)   John gives Mary apples**

**9)   * John gave Mary on Tuesday in the courtyard**

It can thus be seen that agent, affected entity, and goal are *obligatory* roles while time and place are not. Note that generic sentences like 8 do *not* quantify over points in time or locations, not even implicitly (e.g. by assuming default values). Utterance 8 asserts that it is one of John's habits to give Mary apples, without any indication of place or time. In particular, utterance 8 is neither synonymous with ''For *any* given point in time, John will be seen giving Mary apples'' nor with ''For *some* point in time...''. Since the interpretation for the clausal representation enforces, under resolution refutation, a binding for each argument value of a predicate (either referential, existential or universal), we could not represent 8 under the ordered argument scheme without an unwarranted commitment to the values of attributes like ''place'' and ''time'' (and any other attributes encoded as argument positions).

Under the thematic role scheme (axiom system 7a above), on the other hand, we treat all attributes as equally peripheral and dispensable. We could perfectly well drop the roles for agent, affected entity and goal without the resulting set of axioms becoming incoherent. However, any rendering of it in natural language would result in an ill-formed utterance, such as 9. Moreover, the vital distinction between obligatory and optional information is not made, either. We believe that this shows that neither of the representation schemes can be used in pure form.

There are a number of syntactic reasons that convince us that an *intermediate position* must be found. Most substantial among them is the basic syntactic distinction between *complements*, which are obligatory (at least in declarative main sentences; cf. [6]:481) and *adjuncts*, which are *optional*. More specifically, we argue that the former correspond to the semantic category of *participants* in eventualities, and the latter to *circumstances*. The most appropriate way to map such constituents into a logical representation is therefore to use fixed argument positions in a complex *main predicate* for the values derived from complements and the (equally obligatory) governing subject (this is the ordered argument component of the compromise), while all others are represented as *auxiliary predicates* (this is the thematic role component). Main and auxiliary predicates are linked through one additional argument for the eventuality identifier. This intermediate position corresponds materially to Davidson's original approach ([7]). Later objections to it (also by Davidson himself), based primarily on the possibility to report in a non-contradictory manner on impossible situations ([8]:87 ff., [5]:93ff.]), fail to convince us.

For current purposes we may suppose that there are three participants in English, viz. **Agent, Affected-Entity, and Goal**. For a sentence like 6 above, with a prototypical bi-transitive main verb with clear complements, this might give the following set of axioms

```
7b) action(sk-5,give,john,sk-1,mary).
    object(sk-1,apple).location(courtyard1,sk-5).
    time(tuesday1,sk-5). manner(furtive,sk-5).
```

If, in a last step, we define sortal restrictions on the possible values that the arguments can take (e.g. `+animate-ness` for **Agent**) and make sure that for each eventuality there can be only *one* main predicate, the resulting representation scheme amounts to a *case frame representation* for the central propositions of sentences with a sound logical foundation.

In this section we move from the problem of the required *number* of participants in eventualities to the topic of their *semantic characteristics*. The ontological intuition behind the use of a fixed number of argument values is that the fillers of a given thematic role have certain semantic characteristics in common, *irrespective* of what concrete predicate is used to denote the relationship. If we continue to use the thematic roles most commonly assumed to exist, viz. **Agent, Affected-Entity**, and **Goal**, and an appropriate predicate scheme of the following form

```
action(Id,Action,Agent,Aff-Ent,Goal)
```

these thematic roles would allow us then to postulate general inference rules like

$\forall\ I_1, P_1, A_1, E_1, G_1, I_2, P_2, E_2\ \text{action}(I_1, P_1, A_1, E_1, G_1)\ \land\ \text{action}(I_2, P_2, E_1, E_2, G_1)\ \rightarrow\ \exists\ I_3:\ \text{action}(I_3, P_2, A_1, E_2, G_1)$

$\text{with}\ P_1 \neq P_2,\ E_1 \neq E_2,\ I_1 \neq I_2 \neq I_3$

that is, "If Agent $A_1$ performs some Action $P_1$ to Affected Entity $E_1$ with Goal $G_1$, and $E_1$, as Agent, performs some other Action $P_2$ to Affected Entity $E_2$ with the same Goal $G_1$, then Agent $A_1$ effectively himself performs Action $P_2$ to Affected Entity $E_2$ with Goal $G_1$".

It would certainly be desirable to have this kind of completely general inference rules but there is so far very little consensus on the uniform semantic characteristics of different thematic roles that would make such rules possible. Somers, at the end of an exhaustive investigation of the literature ([9]), came to the conclusion that it is *not* possible to find a small and closed set of thematic roles which capture all the semantically relevant information about role fillers. He distilled off four very general *"inner roles"*, modelled (and named) after the prototypical processes of movement, viz. "source", "path", "goal", and "local" (± affected entity), and then defined a number of *parameters* which combine with the inner roles to give concrete participant descriptions. These parameters correspond to very general types of eventualities, of which Somers suggested six, beginning with straightforward movement in space and time ("locative" and "temporal"), over actions (with an active agent: "active") and processes (without an active agent: "objective"), to immaterial changes of (psychological or legal) eventualities ("dative") and finally eventualities without any agent (weather verbs etc.; "ambient"). Consequently, semantic generalisations are allowed only within a given type of eventuality. This explanation of the facts is very attractive from a computational perspective ([10]), in at least two respects.

## 6. A mixed-level representation

**First**, inferential relationships in terms of *types* of eventuality are at least as useful as completely general inference rules operating over true deep cases. Using a predicate scheme of the following form

```
Parameter(Id,Eventuality,Source,Path,Goal,Local)
```

we can formulate rules like

```
∀ I₁,E₁,S₁,P₁,G₁,L₁,I₂,E₂,P₂,G₂:
    locative(I₁,E₁,S₁,P₁,G₁,L₁) ∧
    locative(I₂,E₂,S₁,P₂,G₂,L₁)
            → ∃ I₃,E₃,P₃: locative(I₃,E₃,S₁,P₃,G₂,L₁)
with: E₁ 〉 E₂ 〉 E₃ (non-overlapping temporal sequence)
    ∧ 〚 E₃ 〛 ⊃ 〚 E₁ 〛 ∧ 〚 E₃ 〛 ⊃ 〚 E₂ 〛
```

Provided we know that both "roll" and "fall" are *moving*-actions (parameter *locative*) we can now perform the inference

```
Since the ball rolled from the center of the table
to its edge and fell from there to the floor
it will have moved (in an unspecified manner)
from the center of the table to the floor
```

Significantly, such inferences are on the right level of granularity for the purposes of natural language understanding.

**Second**, it is now relatively easy to recognise the "inner roles" for a given verb phrase as we no longer require that there must be common semantic characteristics to *all* fillers of a given inner role, irrespective of the type of eventuality. We can therefore apply a fairly shallow and straightforward mapping from grammatical structure to inner role structure, using mainly the subcategorisation information for the given verb type. In many cases, the inner roles are linguistically realised by means of particularly obvious *spatial expressions*, even when used for types of eventualities that have nothing to do with space at all, as in "to translate *from ... to*" (probably a reflex of the historical development of language). The mapping may be different for the various types of eventualities but for each given type they must be the same, or else the assignment of roles will be useless for the purpose of inferences. In fact, this observation can be used to inductively infer what types of eventualities should be distinguished ([6]:481): Those verbs whose role frame allows the same inferences denote the same type of eventuality. To actually infer types of eventualities that way, presumably on the basis of corpus material, would require a massive investment in time and resources which, to the best of our knowledge, has never been made. This is why we use, for the time being, a very small subset of eventuality types (after [11] and [12]), viz. "states", "processes" ("actions" where the agent is active), and "events" ("performances" with an active agent). In view of the data presented above it seems fairly clear that a parametrised role concept is much easier to implement than the elusive fully general "deep case" approach (and probably it is more useful in actual applications).

Once we have determined what roles and what eventuality parameters we have to distinguish, and how syntactic structures relate to them, we must decide how many *types of modifiers* should be used, i.e. how many additional expressions of circumstances must be introduced. Expressions of time and location are the most obvious ones and manner, cause, and reason may also be uncontested candidates, but what about the innumerably many other ways to express additional information about events and actions? Are there hard criteria telling us what kind of modification should be dignified with its own role predicate? The linguistic evidence strongly suggest the appropriateness of a *layered representation system*: Instead of trying to find a closed set of modifiers it seems reasonable to claim that there is a small, finite, core set of fairly general (and easy to recognise) modifiers, and an outer layer of (arbitrarily specific) additional circumstance descriptions, whose number is potentially infinite.

The set of core modifiers we use so far consists of just **purpose**, **method**, **tool**, **beneficiary**, and **manner**. For all modifiers that we cannot analyse in these terms we resort to a *lower level of abstraction* and represent them in the logic as *themselves*, i.e. they become non-logical constants. In 10

**10) On Tuesday, John gave Mary a nice computer table against her will**

we do, for instance, not know whether ''computer table'' means ''table next to the computer'', ''table on which to put a computer'', ''table designed by computer'', or any of a number of other possible readings, and it is equally unclear what kinds of semantic relationships are encoded in the prepositional phrase ''against her will''. We therefore turn the unanalysed preposition and the implicit ''of'' in the personal pronoun into non-logical constants, and for the nominal compound we create an artificial constant (`by_with_for`) encoding the unanalysed relationship between its constituent parts. That way we get a *mixed-level representation*, combining expressions on *three levels of abstraction*:

1. **Expressions of level 1:** A fixed number of general, obligatory and unique thematic role fillers
2. **Expressions of level 2:** A fixed number of general but neither obligatory nor unique modifier predicates
3. **Expressions of level 3:** An unlimited number of arbirarily specific, optional and non-unique circumstantial descriptions

An appropriate representation of utterance 10 would therefore be

```
10a) locative(give,sk-1,john,sk-3,mary).
     object(table,sk-3). object(computer,sk-2).
     object(will,sk-4). property(nice,sk-3).
     time(sk-1,tuesday).
     circumstance(by_with_for,sk-3,sk-2).
     circumstance(against,sk-1,sk-4).
     circumstance(of,sk-4,mary).
```

with expressions of level 1 at the top, followed by expressions of levels 2 and 3. By using different levels of semantic granularity in the knowledge representation, we can refine our knowledge in a fully incremental fashion

- We always represent knowledge at the most general level possible at the given point in time but allow for arbitrarily specific entries.
- If we should later discover more general entailments between some of these specific entries, we assert appropriate rules, i.e. *meaning postulates*, found empirically and added incrementally.
- The more information we gather, the denser the network of such entailments will become, without any need to restructure the knowledge base.

In the given application context of passage retrieval, meaning postulates are particularly important to increase recall. Consider a passage

**11) Natural language question answering systems**

where we speak about *systems* that perform certain actions by means of natural language, with the logical representation:

```
property(natural,sk-28)/1/11.
object(system,sk-30)/1/11.
object(language,sk-28)/1/11.
circumstance(by_with_for,sk-30,sk-28)/1/11.
object(question,sk-29)/1/11.
eventuality(answer,sk-31,sk-30,sk-29)/1/11.
```

But consider now the query

**12) Natural language questions**

which is obviously about *questions* phrased in natural language. Nevertheless we would definitely want query 12 to retrieve passage 11. We can increase recall of a retrieval system suitably by using meaning postulates such as

```
circumstance(by_with_for,O1,O2)
  <- eventuality(AType,Ev,Ag,O1),
     circumstance(by_with_for,Ag,O2).
```

which will also allow us to retrieve document 11 through query 12.

## 7. Variable-Depth evaluation

Unfortunately, the use of a mixed-level representation does not come for free. The lack of a known and fixed level of generality must be compensated for by a large number of meaning postulates, and these tend to get very detailed, i.e. their branching factor is high. On top of that, in most application contexts we will have to use inheritance hierarchies. If we use meaning postulates and inheritance relationships *whenever* they are applicable we will run into insurmountable problems with the size of the search space. We need a mechanism that controls the search procedure suitably. This is where *variable-depth evaluation* becomes useful ([13], [14]).

We can distinguish two types of criteria that may control the search procedure: *External criteria* (such as the maximum number of passages to be retrieved), and *internal criteria* determining the resources to be used for a given proof step. External criteria are straightforward to implement: If a relatively shallow level of evaluation has already produced a large number of results (e.g. relevant passages), it is better to stop the proof at this level and present the results to the user who might then be able to re-phrase the original query and make it more specific. External criteria stop the proof when it is, as it were, *too* successful. Internal criteria influence the proof process when it is *not enough* successful, and steer it towards more promising branches of the search space. The single most important internal criteria are probably the *total number of inferences* allowed for the proof of a given term, the *type of the rules* to be used, and the *weights of individual rules* (indicating, for instance, their reliability or general usefulness).

In LogDoc, we use a combination of external and internal criteria. The different computational costs of using various types of rules, and also their different usefulness as determined by preliminary experiments, suggested the following strategy:

1. if we have found more than M passages using only *direct* matches of thematic role relations, we do not use any of the meaning postulates

2. if we have found fewer than N (N < M) passages we begin to use meaning postulates, first those on level 2 (defining modifiers), then those on level 3 (defining circumstantial descriptions)

3. if we have found fewer than O (O < N) passages so far, we also try inheritance hierarchies

We found that a reasonable weighting of these values is M:N:O = 3:2:1. The absolute threshold values used so far are very low (M=15, N=10, O=5), due to the very small size of the sample of documents used so far. However, we seem to get a reasonable ordering of passages.

## 8. Implementation and results

LogDoc is implemented as a fully functional prototype (on a previous version, see [15]). It uses a breadth-first, bottom-up chart parser over a very small grammar of English (represented as Definite Clause Grammar rules). The translation into logic is performed in a compositional, Montagovian manner, i.e. exactly following the syntax structures. Syntactic ambiguities are dynamically pruned on the basis of two weighted general preference rules, the strong Principle of Right Association and the slightly weaker Principle of Minimal Attachment, modified by local syntactic and lexical semantic criteria, each with empirically determined preference weights, and by absolute frequencies of lexical items. Surviving ambiguities are then filtered on the basis of the proportional distance in preferences between ambiguous readings, after which logically equivalent readings are collapsed. These steps are, in the majority of cases, sufficient to suppress all, and only, the intuitively unavailable readings of ambiguous constructions. Ambiguities that still survive are added as disjunctive terms to the logical data base. This required a suitable extension of the syntax of HCL, and the proof procedure had also to be modified correspondingly. When a phrase cannot be analysed by the parser in its entirety, the maximal fragments are translated into logic individually, and an attempt is made to prove these logical expressions. If a proof fails over the axioms derived from the same sentence, the system incrementally relaxes proof criteria. It first tries to prove the complete query over axioms derived from the same document and, failing that, it gradually breaks the query into ever smaller pieces, according to the importance of different types of terms and the links between them (links with circumstantial descriptions are broken first, those between thematic role fillers are kept longest). This makes the system remarkably robust: Whenever a linguistic proof is impossible, LogDoc gradually falls back to, ultimately, the standard IR regime.

As test documents the system uses only a few dozen abstracts in English from the field of NLP, and the grammar is also very small. It is therefore much too early to conduct any meaningful experiments to compare the performance of LogDoc with that of other systems. However, the basic principles, viz. high initial precision, extensibility, robustness, and the use of variable-depth inference to increase recall without loss of precision and with acceptable run-time behaviour, can be shown to work as designed. It is certain that considerable work will be needed to make the system perform comparably well for larger grammars and larger samples of text.

## 9. Related work

The idea that a meaning representation language could be used as an indexing and retrieval language is not entirely new. There are a few systems that use a (very limited) amount of logic for retrieval (e.g. [16]). However, then notion of variable depth representation is less widespread. One system that is, on the surface, very similar to LogDoc in the way it uses logic to encode knowledge on different levels of generality, is SILOL by Sembok and Rijsbergen ([17]). Since the semantic import of many syntactic relationships is either ambiguous or altogether unclear, Sembok and Rijsbergen introduce what they call ''generalised relationships''. These relationships should cover the common components in such cases of ambiguity or vagueness, and therefore correspond to the unanalysed lexical items turned into logical constants in our system. However, Sembok and Rijsbergen then generalise this procedure and use it for *all* types of syntactic relationships, even the ones which are straightforward to analyse. The relationship between subjects and verbs becomes `sv(X,Y)`, that between subject, verb and direct object `vso(X,Y,Z)` etc., and *all* prepositions are treated as generalised relationships in their own right. That way the logical representation encodes merely the syntactic structures themselves, not even part of their meaning. But without a logical model the whole problem of syntactic variability is back with a vengeance. An approach that is closer to ours is that of the system TACITUS ([18]). Hobbs and Stickel represent unanalysable syntactic constructions (such as nominal compounds) by logical predicates (e.g. `nn`) and add axioms for the most common possible relations (for nominal compounds, *part*, *sample*, *for*). However, they do not seem to use variable-depth evaluation techniques to limit and guide search.

**Acknowledgment**

This work has been funded by the Swiss National Science Foundation (projects 4023-026996 and 1214-045448.95/1)